# Computational Metacognition

**Michael T. Cox**  MICHAEL.COX@WRIGHT.EDU
**Zahiduddin Mohammad**  MOHAMMAD.48@WRIGHT.EDU
**Sravya Kondrakunta**  KONDRAKUNTA.2@WRIGHT.EDU
**Ventaksampath Raja Gogineni**  GOGINENI.14@WRIGHT.EDU
Computer Science and Engineering, Wright State University, Dayton, OH 45435 USA

**Dustin Dannenhauer**  DUSTIN.DANNENHAUER@PARALLAXRESEARCH.ORG
**Othalia Larue**  OTHALIA.LARUE@PARALLAXRESEARCH.ORG
Autonomy Research Group, Parallax Advanced Research, Beavercreek, OH 45431 USA

## Abstract

Computational metacognition represents a cognitive systems perspective on high-order reasoning in integrated artificial systems that seeks to leverage ideas from human metacognition and from metareasoning approaches in artificial intelligence. The key characteristic is to declaratively represent and then monitor traces of cognitive activity in an intelligent system in order to manage the performance of cognition itself. Improvements in cognition then lead to improvements in behavior and thus performance. We illustrate these concepts with an agent implementation in a cognitive architecture called MIDCA and show the value of metacognition in problem-solving. The results illustrate how computational metacognition improves performance by changing cognition through meta-level goal operations and learning.

## 1. Introduction

The *computational metacognition* process is analogous to an action-perception cycle in an intelligent agent (Cox, 2005). But, instead of perceiving the environment and acting in the world, metacognition monitors cognition and acts to control cognitive activity. In humans, such introspective processes can prove beneficial, as in a student's proper regulation of classroom learning (Price, Hertzog, & Dunlosky, 2009) or in a game show contestant's judgment of knowing answers to questions posed by the show's host (Reder & Ritter, 1992). Yet given the significant overhead and complexity it exhibits (Conitzer, 2011), metacognition is not a panacea for all computational tasks nor is it always beneficial in human performance (Norman, 2020; Wilson & Schooler, 1991). In this paper, we will examine the kinds of metacognitive activities that can improve a cognitive system's performance, and we will demonstrate these concepts in an implemented problem-solving domain.

In some sense, metacognition is an add-on to a cognitive system. If a cognitive system has a certain behavior, then a metacognitive system should be able to observe that behavior and improve the performance of that system by changing its behavior given some parameters under its control. But here we are speaking of cognitive behavior (i.e., problem solving) rather than the physical behavior of the system in the world. The impact on the actual behavior of the joint system is indirect and improves performance by improving thinking. As such, metacognition is one of the characteristics of intelligence that separates humans from mere reinforcement machines. This paper presents a computational approach to metacognition that seeks to clarify the manner in which such indirect levers can exert such influential effects on performance. Its contribution is to specify the mechanism of metacognition in computational terms within the context of an existing cognitive architecture and to evaluate these concepts in planning problems. Previous work has discussed the process of detecting metacognitive expectation failures; the paper here discusses the processes of meta-level explanation and goal generation as well as meta-level planning and learning. This is our first fully complete implementation of the metacognitive cycle.

Section 2 describes our concept of computational metacognition consisting of explanatory, immediate and anticipatory metacognition. Subsequently, Section 3 shows how a cognitive architecture instantiates many of these ideas computationally and illustrates the principles with examples. Section 4 then evaluates this approach empirically in a simple, plant-protection domain. Finally, Section 5 enumerates related research, and Section 6 summarizes and concludes.

## 2. Computational Metacognition

Though the concept of computational metacognition has many variations in the literature, having both broad and narrow interpretations, virtually all theories divide it into some kind of *introspective monitoring* and *meta-level control*. In the broader sense, any self-directed process can be included under its umbrella; whereas, the stricter definition of meta-x as "x about x" (see Hayes-Roth, Waterman, & Lenat, 1983) constrains the subject to cognition about cognition. This more narrow definition excludes related concepts such as meta-knowledge (i.e., knowledge about knowledge) which is not a cognitive process per se (Cox, 2011). But broadly speaking in all its interpretations across the literature, metacognition is surely "the many-headed monster of obscure parentage" of which Brown (1987) speaks; it is not a single monolithic construct. Instead, we claim that metacognition has three fundamental forms (see Table 1).

*Table 1*. Types of Computational Metacognition

| **Explanatory** | **Immediate** | **Anticipatory** |
|---|---|---|
| Past | Present | Future |
| Hindsight | Insight | Foresight |
| Retrospective | Introspective | Predictive |

*Explanatory metacognition* is a reflective process triggered by failures in previous cognitive operations and thus represents a process akin to hindsight. *Immediate metacognition* represents introspective run-time control of cognition, analogous to physical eye-hand coordination.[1] *Anticipatory metacognition* is a reflective judgement of future cognitive performance and hence represents self-directed foresight. This paper has little room to discuss all three forms. Instead, it will focus on the explanatory category of metacognition and briefly examine the anticipatory one.

---

[1] This is also related to decision-theoretic metareasoning applied to partial computations (Horvitz, 1990) and to bounded rationality decisions for anytime-planning (Zilberstein, 2011). In contrast, our approach is symbolic and non-statistical.

## 2.1 Explanatory Metacognition

*Expectations* play important functional roles in cognition such as the monitoring of plan execution (Dannenhauer & Munoz-Avila, 2015; Pettersson, 2005; Schank, 1982), managing comprehension, especially natural language understanding (Schank & Owens, 1987), and influencing emotion (Langley, 2017). Expectations are knowledge artifacts that enable cognitive systems to verify their behavior is working as intended. The agent checks if discrepancies exist between its expectations and its observations of the state of the world. When such a discrepancy is detected, an *expectation failure* is said to occur. Different means of addressing expectation failures have been proposed including plan adaptation (i.e., modifying a plan to be executed) (Munoz-Avila & Cox, 2008), learning (i.e., acquiring a new piece of knowledge) (Munoz-Avila, 2018; Ram & Leake, 1995a), and goal reasoning (i.e., changing the goals pursued or formulating new goals to be achieved) (Aha, 2018; Roberts, Borrajo, Cox, & Yorke-Smith, 2018). Expectation failures are signals that a problem may have arisen. Explanations of the problem then provide the basis for goal formulation and thus facilitate problem-solving. Note that we are referring to a kind of internal explanation or self-diagnosis process rather than an external explanation to another agent.

Similarly, *metacognitive expectations* (Dannenhauer, Cox, & Munoz-Avila, 2018) play an analogous role to the expectations discussed above. But here the expectation concerns the outcome of cognitive processes rather than events in the world. As such, it relates the mental states immediately before and resulting from a specific mental process or action. An individual *mental state* $s^M = (v_1, \dots, v_n)$ is a vector of variables; whereas, a *mental action* $\alpha^M$ performs reads from or updates to variables in a mental state. The metacognitive expectation then is represented as the triple $(s_i^M, \alpha_i^M, s_{i+1}^M)$ where $\alpha_i^M$ is the current cognitive process and each $s^M$ represents the system's mental state immediately preceding and following from it.[2] That is, for a given cognitive process, the expectation specifies constraints on memory associated with the execution of the process.

For example, one might expect commitment to a current goal before a planning process and a plan to be in memory after planning finishes. That is, the expectation would be represented as the triple $(g_c \in s_i^M, Plan, \pi \in s_{i+1}^M)$. If a plan does not result after planning executes, then something went wrong in cognition (Cox & Dannenhauer, 2016). The reasoning from such metacognitive expectation failures is in hindsight, because it happens after a mistake occurs. The function of metacognition is to explain what caused the cognitive-level failure and to formulate a meta-level goal to mitigate the problem. For example, changing the goal minimally to something easier but serving the same end may enable successful planning when physical resources are scarce (see Cox & Veloso, 1998).

Multiple *meta-level goals* are available to affect cognition. Currently, we have implemented three different meta-level goal types. The first aims to change cognition directly, while the other two represents indirect change.

1. To change the reasoning method (e.g., change from state-space planning to case-based planning) (Cox & Dannenhauer, 2016);
2. To change cognition by changing the goals of the system (Cox, Dannenhauer, & Kondrakunta, 2017);
3. To change cognition by learning new knowledge structures (Mohammad, 2021; Mohammad, Cox, & Molineaux, 2020).

---

[2] Dannenhauer, et al. (2018) defined metacognitive expectations as a Boolean function taking such a triple as input. But, the differences are unimportant for the purposes of this paper.

This paper will focus on the third type, that is, *learning-goals* (i.e., an explicit meta-level goal to learn a specific piece of knowledge) (Cox, 1997; Cox & Ram, 1999; Ram & Leake, 1995b) as a metacognitive response, but the others are equally as important. Plans can be generated at the meta-level composed of "actions" such as performing learning in pursuit of a learning goal or executing different *goal operations* (Cox, et al., 2017; Kondrakunta, Gogineni, & Cox, in press) such as goal change in pursuit of the second type above. But, the goal delegation operation has a particular role in anticipatory metacognition which spans both cognition and metacognition.

**2.2 Anticipatory Metacognition**

The concept of anticipatory metacognition is forward looking. Humans demonstrate foresight concerning their cognitive prowess or the lack thereof on a regular basis. Likewise, cognitive systems can benefit from an ability to predict whether or not they can achieve their goals as opposed to exhaustively trying all possible solutions before acquiescing. Unlike *anticipatory thinking* (Amos-Binks & Dannenhauer, 2019) that predicts plan execution failures and seeks to mitigate the vulnerability of a plan after it is generated, our approach is to anticipate a failure given the goal but before planning is performed. Making such a prediction, an agent can simply delegate some of its goals to another agent willing to help it and thus mitigate the amount of work to be done.

In contrast to explanatory metacognition, which is triggered by metacognitive expectation failures, anticipatory metacognition is triggered by the presence of *suspended goals*. Suspended goals are ones that were part of an agent's current goal set but were determined to be problematic given physical resource limitations for example. This condition suggests that planning or other reasoning will fail in the future. As a result, metacognition can generate a meta-level goal to change the cognitive-level goal. In this case, it is a change from the agent's goal to another agent's goal. Thus, the *goal delegation operation* is partially performed at the cognitive level and partially at the meta-level. Suspending the goal is cognitive; whereas, reasoning about another agent's knowledge, skills and goals in relation to the agent's own goals is associated with Theory of Mind (Goldman, 2006; Gopnik, 2012; Wellman, 1990) and with metacognition. This allows the system to target a good candidate agent for the delegation. Finally, if the agent decides to delegate one or more of its goals and determines a candidate to achieve them, the agent still needs to make an actual request, explain the need for the request, and negotiate the favor. Such reasoning is once again situated at the cognitive level and in the speech acts (Searle, 1969) executed in the world. Gogineni, Kondrakunta, and Cox (in press) provide a complete example of this hybrid process with experimental evidence supporting the benefits of this type of metacognitive activity.

Given space limitations, this paper cannot but sketch the mechanisms behind anticipatory metacognition. Instead, we will describe the detailed process of explanatory metacognition and show how it is implemented in a particular cognitive architecture. Specifically, we will explain how it manifests as a combination of introspective monitoring and meta-level control. With these details in hand, we will then provide an empirical evaluation that demonstrates the benefit of computational metacognition for cognitive systems.

**3. Architectural Structure and Implementation**

The *metacognitive integrated dual-cycle architecture (MIDCA)* is a cognitive architecture that models both cognition and metacognition for intelligent agents (Cox et al., 2016; Cox, Oates, & Perlis, 2011; Paisner, Cox, Maynord, & Perlis, 2014) and focusses upon various goal operations including goal change, goal formulation and goal delegation (Cox & Dannenhauer, 2017; Cox, et al., 2017).³ It consists of "action-perception" cycles at both the cognitive level and the metacognitive level (see Figure 1). In general, a cycle performs problem-solving to achieve its goals and then comprehends the results of its behavior and interactions with other agents in its environment. The problem-solving portion of each cycle consists of three phases: intention; planning; and the action-execution/control phase. The comprehension portion consists of perception/monitoring, interpretation, and goal evaluation.

The representations in MIDCA and our formal notation borrow much from the AI planning community (e.g., Ghallab, Nau, Traverso, 2004). We rely on the notion of a *state transition system* $\Sigma = (S, A, \gamma)$ for representing a planning problem. In $\Sigma$, $S$ is the set of all possible world states, $A$ is the set of actions the agent can take, and $\gamma$ is the successor function $\gamma: S \times A \dashrightarrow S$. A planning problem is represented as $\mathcal{P} = (\Sigma, s_0, g)$ where the initial state $s_0 \in S$ and the goal $g \in G \subset S$. A plan $\pi = \langle \alpha_1, \alpha_2, \ldots, \alpha_n \rangle$ is a sequence of actions $\alpha_i \in A$. The plan $\pi$ represents a solution to $\mathcal{P}$ by achieving $g$ if and only if its iterative execution from $s_0$ results in a final state $\gamma(s_{n-1}, \alpha_n)$ that entails the goal expression. Note however, that a problem need not always start from some arbitrary initial state but instead may arise during the planning or the plan execution related to some previous problem (Cox, 2020).⁴

At the cognitive level, comprehension starts with observations in terms of percepts ($\vec{p}$) of the world ($\Psi$) whereby the Perceive phase infers the objects in the environment and the relationships between them. The Interpret phase takes as input the resulting relational state ($s_j$) and the expectations in memory to determine whether it is making sufficient progress. It is here that a model of the world ($M_\Psi$) is inferred from its observations, and new goals ($g_n$) are generated when the model indicates problems or opportunities. Interpret adds any new goals to the goal agenda ($\hat{G} = \{g_1, g_2, \ldots g_n\}$). The Evaluate phase incorporates the concepts inferred from Interpret and checks whether the current goal set ($g_c \in \hat{G}$) is achieved, i.e., whether $s_j$ entails $g_c$. If so, $g_c$ is removed from the agenda and set to the empty set (i.e., $\hat{G} \leftarrow \hat{G}/g_c$ and $g_c \leftarrow \{\}$).

In cognitive-level problem solving, the Intend phase commits to a new goal set $g_c$ from those available in $\hat{G}$ if the old goal set is empty. The Plan phase then generates a sequence of actions (i.e., the current plan, $\pi_k = \langle \alpha_1, \alpha_2, \ldots \alpha_n \rangle$) to perform in pursuit of its goals. The Act phase executes the steps of the plan ($\alpha_i$) one at a time to change the environment through their effects. MIDCA will then use expectations about these actions in subsequent cycles to evaluate the execution of the plan. At the end of each cognitive phase, MIDCA performs an entire metacognitive cycle as explained below.

Like cognition, MIDCA partitions metacognition into introspective monitoring (analogous to cognitive-level comprehension) and meta-level control (analogous to problem-solving). Introspective monitoring detects metacognitive expectation failures and formulates goals to change cognition in response. Meta-level control generates and executes plans to achieve these goals and thus improve cognitive performance. See Dannenhauer, et al. (2018) for a more formal specification of the two-cycle MIDCA mechanism. Here we give enough detail to capture the metacognitive process computationally and allow the reader to follow the examples and understand the evaluation.

---

³ MIDCA version 1.5 is open-source and runs on python 3.8. The source code and documentation are publicly available at https://github.com/COLAB2/midca. See also www.midca-arch.org.

⁴We actually assume a somewhat different problem representation $\mathcal{P}_c = (s_c, s_e, Bk, H_c)$ where $s_c$ and $s_e$ represent the currently observed and expected world states, $Bk$ is the background knowledge and $H_c$ is the episodic problem-solving history of the agent (see Cox, 2020). But the differences are not important for the purposes of this paper and may be ignored by the reader.

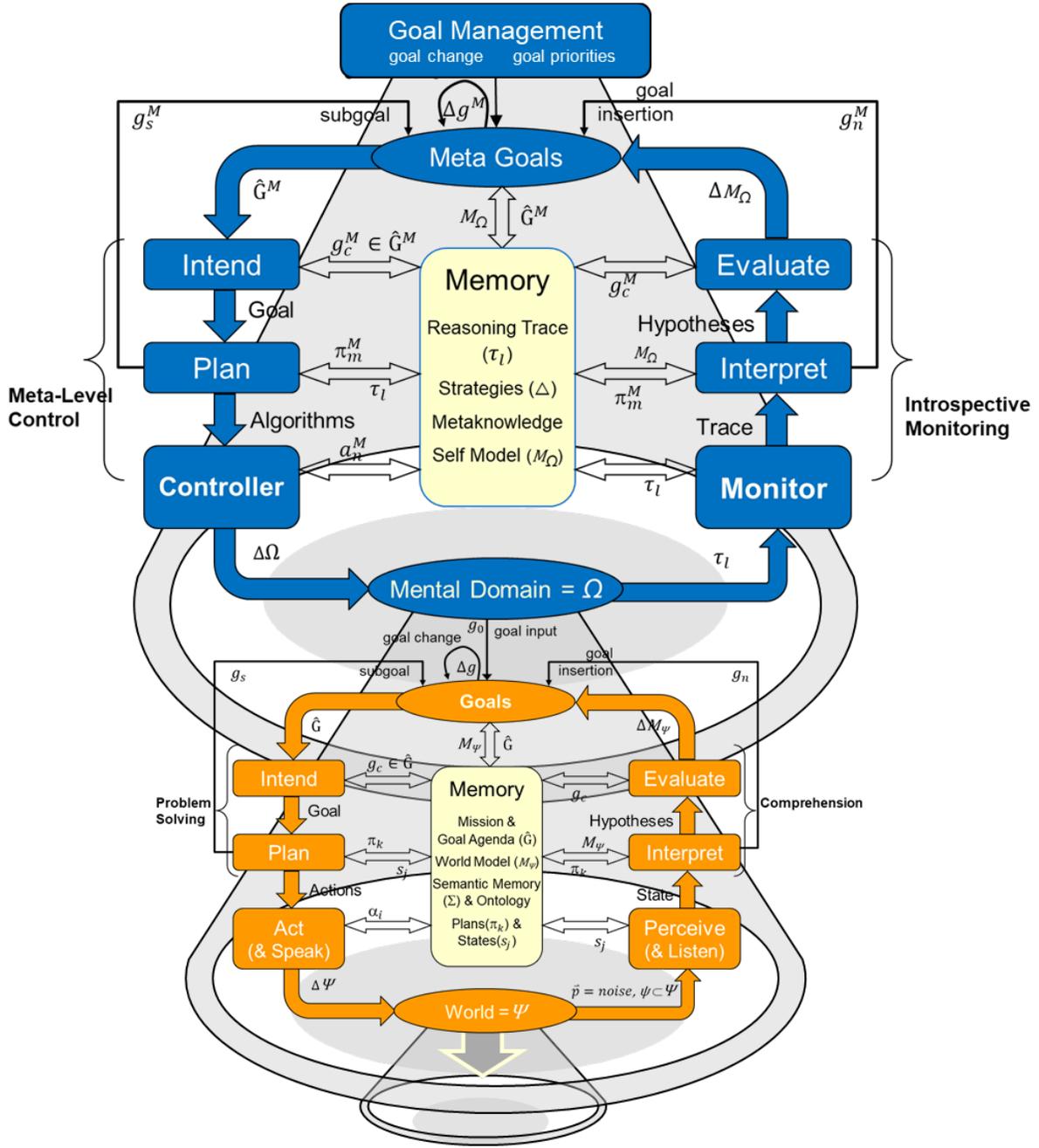

*Figure 1.* The metacognitive integrated dual-cycle architecture and the flow of knowledge between computational phases. The lower (orange) cycle represents cognition, receiving stimuli in the form of percepts and acting upon the environment, thus changing the world state. The upper (blue) cycle represents metacognition, receiving an introspective trace of cognition and controlling the cognitive level through goal operations and learning. Note that all knowledge structures $k$ that are metacognitive have the superscript $k^M$.

### 3.1 Introspective Monitoring

At the meta-level, introspective monitoring starts with a *trace* ($\tau_l$) of the mental domain, that is, of the activity at the cognitive level (again see Figure 1). In support of this knowledge structure, Dannenhauer, et al., (2018) defines an agent's *self-model* (i.e., a model of the cognitive level) as $\Omega = (S^M, A^M, \omega)$ where each $s_i^M \in S^M$ is a possible mental state, each $\alpha_i^M \in A^M$ is a mental action, and $\omega$ is a cognitive transition function.[5] In MIDCA, a mental state is represented as a vector of length $n = 7$ where $s_i^M = (g_c, \hat{G}, \pi_k, M_\Psi, D, E, \alpha_i)$. That is, it consists of the current goal, the goal agenda, the current plan, the current world state, discrepancies detected, explanations generated, and the last action executed in the world. MIDCA employs the following mental actions: Perceive, Detect Discrepancies, Explanation, Goal Insertion (i.e., updates the set of goals $\hat{G}$ to be achieved), Evaluate, Intend, Plan, and Act. Each of these represent either one of the six phases at the cognitive level or a subprocess within Interpret. The *cognitive transition function* is defined as $\omega: S^M \times A^M \longrightarrow S^M$. That is, given a current mental state, $\omega$ provides the successive mental state resulting from the execution of that mental action and hence is a source of metacognitive expectations. Finally, the trace is an interleaved sequence of mental states and mental actions. More specifically, $\tau_l = \langle s_0^M, \alpha_1^M, s_1^M, \alpha_2^M, \ldots, \alpha_n^M, s_n^M \rangle$ where $s_0^M, \ldots, s_n^M \in S^M$ and $\alpha_1^M, \ldots, \alpha_n^M \in A^M$.

The Monitor phase takes the recorded trace and places it in memory for the meta-level Interpret phase to examine. As with the cognitive level phase, meta-level Interpret seeks to detect and explain discrepancies between specific metacognitive expectations and its "observations" in segments of the cognitive trace. As mentioned in Section 2.1 these expectations consist of the memory states before and after a given cognitive phase or phase component. If a discrepancy exists, then Interpret attempts to explain the discrepancy and formulate a meta-level goal.

Consider an agent that is learning to take care of a garden having goals such as the preservation of native plants and the removal of invasive ones. Its knowledge of plant care, such as the correct application of herbicide, may be flawed, and therefore it may make mistakes in both the execution of solutions (i.e., its behavior in the world) and in the derivation of those solutions (i.e., its reasoning to solve problems). For example, the agent's model of the spray action may lack the knowledge that spraying herbicide will kill plants in locations adjacent to those it targets. When the goal of native plant preservation is violated after spraying invasive ones close to the natives, a normal discrepancy occurs. However, when the Interpret phase fails to explain this discrepancy, a metacognitive expectation fails. That is, the cognitive trace $\tau_l$ does not adhere to the expectation ($discrepancy \in s_i^M$, Explain, $explanation \in s_{i+1}^M$) since the Explain component of the Interpret phase failed to produce the expected explanation.[6] Now the discrepancy at the metalevel must be explained.

At the meta-level Interpret phase, the idea is to detect discrepancies, explain what caused the discrepancy, and then generate a goal to remove the cause. Metacognitive explanation is a case-based process that reuses old explanations from a library of *meta-explanation patterns (Meta-XPs)* in a similar manner to Cox and Ram (1999; Ram & Cox, 1994). Figure 2 shows a Meta-XP applied to the gardening example. MIDCA retrieves an explanation and then checks for applicability by inspecting the Meta-XP's pre-XP nodes. These sink nodes in the graph of the knowledge structure represent those conditions that need to hold for the XP to be relevant. A distinguished node among them called the Explains node is the concept being explained. As such, the XP provides a causal chain from the XP-asserted nodes (i.e., the XP's antecedents) to the explains node. In the running example, the Meta-XP of Figure 2 explains why the cognitive level Interpret phase failed to generate an explanation when MIDCA was currently at the Interpret phase and a native-plant was not preserved. If that is the case, then the XP says that this is caused when the spray action was executed, no other agent exists nearby, and knowledge of the spray action is incomplete.

---

[5] The self-model $\Omega$ is similar to the state transition system $\Sigma$ defined previously but is a meta-level knowledge structure.
[6] Note that if another gardener was observed concurrently spraying herbicide on the adjacent native plant, then an explanation as to why the native plant was killed could be produced and a goal to stop the gardener from repeating that in the future formulated. If this actually was the case, then no metacognitive expectation failure would occur.

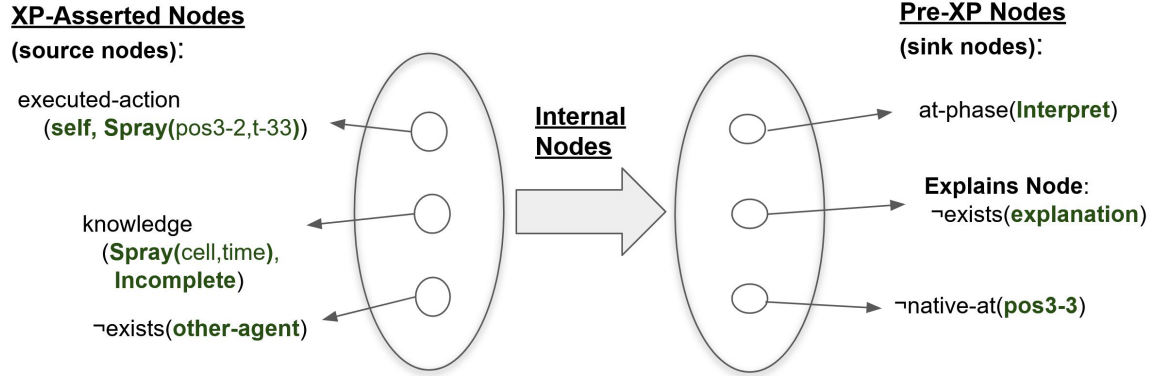

*Figure 2.* Meta-explanation pattern for failed action execution due to a poor action model. An explanation pattern (Schank, 1986) is a causal knowledge structure representing prior experience. It maps the causes of the explains node (i.e., what is being explained) from the antecedents (i.e., the XP-asserted nodes) through the internal nodes to the consequents (i.e., the pre-XP nodes).

The meta-level Interpret phase then takes this knowledge structure and generates a goal from the set of XP-asserted nodes. Here the goal is to negate the middle node, that is, to make the knowledge of the spray operator not incomplete (i.e., to learn a better action model). The specific form output to the meta-level goal agenda is (learned spray, $s_{i+1}$). See Gogineni, Kondrakunta, Molineaux, & Cox (2018) for further details concerning XP retrieval, selection and application at the cognitive level. The same techniques are used at the meta-level.

### 3.2 Meta-level Control

Like problem-solving at the cognitive level, meta-level control consists of an Intend phase, a meta-level Plan phase, and a Controller. Intend takes the meta-level goal agenda ($\hat{G}^M$) and chooses a subset as the current meta-level goal set ($g_c^M$). Details are available in Kondrakunta and Cox (in press; Kondrakunta, 2017) for the decision mechanism. In our running example, only the single learning goal exists in the agenda, and so Intend selects it as the current goal set.

The Plan phase of the meta-level then takes the current goals and generates a plan $\pi_m^M$ to achieve it using the *fast-downward stone soup* planner (Helmert, 2006; Helmert, Roger, & Karpas, 2011). Unlike normal planners that generate sequences of actions in the world to achieve environmental states, MIDCA uses fast downward to achieve meta-level goals. Given the learning goal (learned spray $s_{i+1}$), Fast Downward must generate a sequence of learning steps that achieves the goal if executed. To perform this task, MIDCA has a number of operators represented in the *Planning Domain Definition Language (PDDL) 2.2* (Edelkamp & Hoffmann, 2004), one of which is shown in Table 2. To select the `perform-learning` action, the agent checks whether it has a

*Table 2*. Action model of the meta-level operator `perform-learning`. Specified in PDDL 2.2, this primitive, planning operator achieves the learning goal (`learned ?op ?current-state`).

```
(:action perform-learning
    :parameters (?op - operator ?current-state - state)
    :precondition
    (and
        (has-discrepancy ?current-state)
        (outdated ?op)
        (caused_discrepancy ?op ))
    :effect
    (and
        (learned ?op ?current-state))
)
```

discrepancy in the current state, it has an outdated operator, and the same operator caused the discrepancy. If all the preconditions are met, then it results in the learning goal.

The Controller then attempts to execute the plan one step $a_n^M$ at a time. In the example, the plan has a single step to perform learning. MIDCA uses the *First-Order Inductive Learner (FOIL)* (Quinlan, 1990) for this paper. FOIL induces function-free Horn clauses from a set of positive and negative concept examples and some background knowledge represented as a set of first-order logical predicates. It performs hill climbing with an information theoretical function and generates a rule that covers all the examples.

The FOIL algorithm is given positive or negative examples experienced during the execution of failed plans. For example, the spray action may have killed a plant in a location directly north of the intended cell. If so, FOIL will learn the following general Horn clause.

spray (pos1, time2) :- spray (pos0, time1), adj_time (time2, time1), adj_north (pos0, pos1)

That is, if the agent sprays a location at time t1, it will be as if it sprays the adjacent location due north at t2. Doing so will then kill all plants in the cell north of the spray. MIDCA then compares such rules generated by the learning procedure with the agent's knowledge of the operator. As a result, it changes the spray operator by adding conditional effects.

```
(forall (?pos - mapgrid)
  (when (and (adj_north ?to ?pos))
        (and (not (native-at ?pos))(not (invasive-at ?pos)) )))
```

After the agent's memory is updated with the modified operator, the metacognitive cycle ends. Once metacognition completes, MIDCA continues with the subsequent cognitive phase. Given an incrementally improved operator as with our example or otherwise with some other positive change to cognition, performance will improve. The following section examines this claim empirically and shows specific benefits to cognitive systems that reason about themselves.

## 4. Empirical Evaluation

We evaluate the claim that metacognition can improve the performance of cognitive systems with a relatively simple planning task. The experiments we perform use the plant protection domain developed at the Naval Research Laboratory.

### 4.1 The Plant Protection Domain

The plant protection domain consists of harmful invasive plants, endangered native plants, a human supervisor, and an agent which navigates to a target cell and deploys herbicides (Boggs, Dannenhauer, Floyd, & Aha, 2018). When an agent deploys herbicides, it will affect the neighboring cells as well. The world in which agents act is a map grid of size 10✕10, as shown in Figure 3. It consists of tiles where a plant occupies a single tile, and the tiles contain at most one plant. Plants are static fixtures that cannot be moved or replanted. The garden area where plants exist lies between (2,2) and (7,7), i.e., the white rectangle. In this domain, two key action models exist: a move operator and a spray operator. The walkway around the planting area exists for the ease of agent movement.

Figure 3 visually depicts a very simple problem with a gardening agent at location (0,0). The goals of the agent are to preserve all native plants and to remove all invasive ones, i.e., native-at(pos3-3) ∧ ¬invasive-at(pos2-5) ∧ ¬invasive-at(pos3-2). Here, the Perceive phase of the cognitive cycle inputs the requested expression for the goals, Interpret adds these goals to the goal agenda, and Intend makes all goals current. Finally, the Plan phase takes the current goal set and generates a plan to move to location (3,2), spray the invasive plant, move to location (2,5), and spray again. Given that the native plant already exists at location (3,3) in the initial state, nothing needs to be done to achieve the goal to have a native plant at (3,3). However, the agent does not realize that spraying in one cell will kill plants in all adjacent cells. Thus, the native plant in (3,3) is also killed.

### 4.2 Experimental Design

Two experiments provide a baseline for determining how metacognition with learning affects the performance of an agent in the plant protection domain. Our tests were run in the 10 × 10 world (as discussed above) with a standard planning agent and a metacognitive learning agent. The standard agent does not perform goal reasoning or learning. It just executes plans to remove all invasives. The learning agent learns a more accurate model of its actions over the range of goals given and can reject harmful goals accordingly. We randomly place invasive and native plants in the map grid to form problems of increasing difficulty (i.e., larger amounts of invasive plants and hence more goals to solve in a problem).

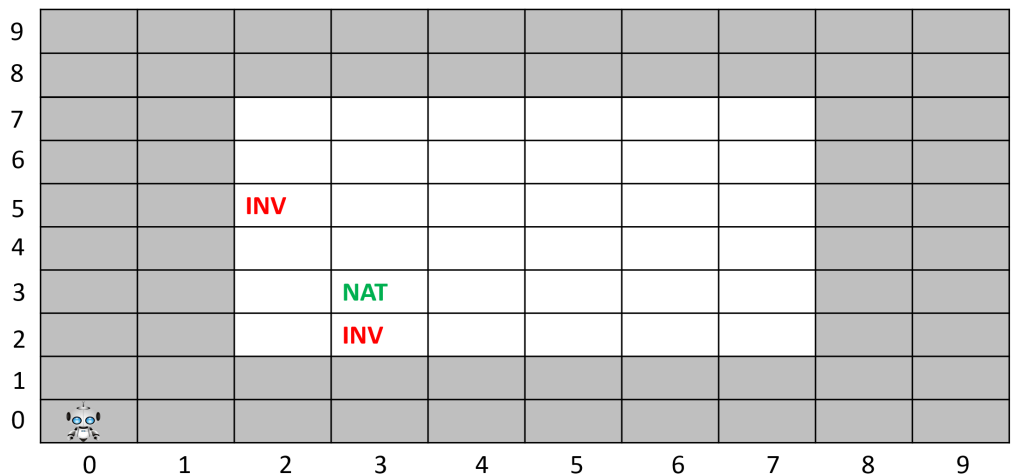

*Figure 3*. A simple problem in the plant protection domain. The central rectangular area in white indicates the garden with native and invasive plants represented by the symbols NAT and INV respectively. The gardening agent at the origin needs to remove invasives and preserve native plants.

For the first experiment, we set the ratio of native to invasive plants at 75:25 and then in another set of trials we set it at 60:40. We also ensure that no two plants are in the same map grid and that all plants are situated between location (2,2) and location (7,7). Experiments were carried out by varying problems with the number of goals ranging from 1 to 20. At each fixed number of goals, we generate 100 random trials thus leading to a total of 2,000 random trials for each experiment. The density of the invasive and native plants on the grid changes in every set of trials as the number of goals increase. Initial goals for this domain will be to remove all the invasive plants and preserve all native plants.

### 4.3 Empirical Results

In Figure 4, the blue line represents the learning agent, and the green line represents the standard agent. From these results, it is clear that the learning agent performs significantly better than the standard agent because a standard agent removes all invasive plants regardless of any plants in adjacent cells. The learning agent outperforms because, after a given number of trials, it improves its use of herbicide. As the spray operator is gradually fixed, the learning agent generates better plans to remove invasive plants while preserving the native ones.

For two different points along the x-axis of Figure 4, Figures 5 and 6 show further details using box plots. A box plot is a standardized way of displaying data such as minimum, first quartile, median, third quartile and maximum. The box is drawn from first quartile to third quartile. The lower line below the box is the minimum value (first quartile value -1.5 * interquartile range) and the upper line is the maximum value (first quartile value +1.5 * interquartile range).

M. T. COX, Z. MOHAMMAD, S. KONDRAKUNTA, V. GOGINENI, D. DANNENHAUER, AND O. LARUE

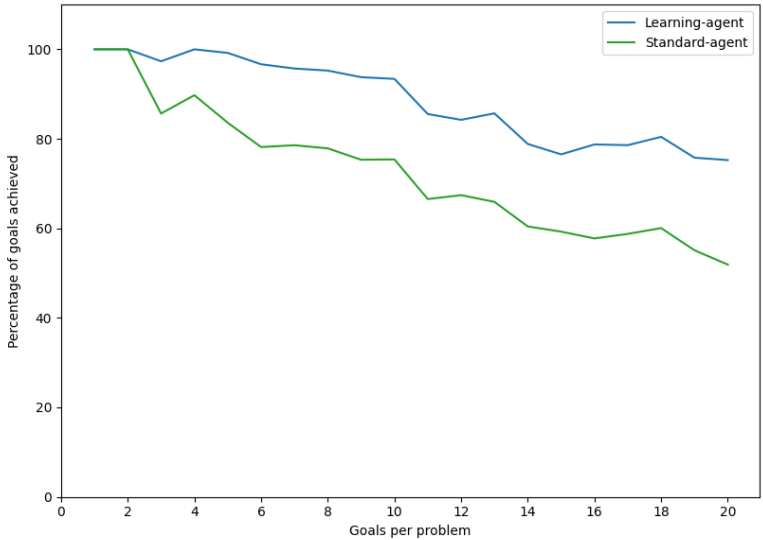

*Figure 4*. Experiment 1 performance as a function of problem complexity. The ratio of native to invasive plants is 75:25 in each of the 2,000 trials. As problems increase in complexity given ever more goals to achieve, performance decreases as a percentage of goals achieved. But, the metacognitive agent that learns a better action model outperforms a standard planning agent that uses no metacognition or learning.

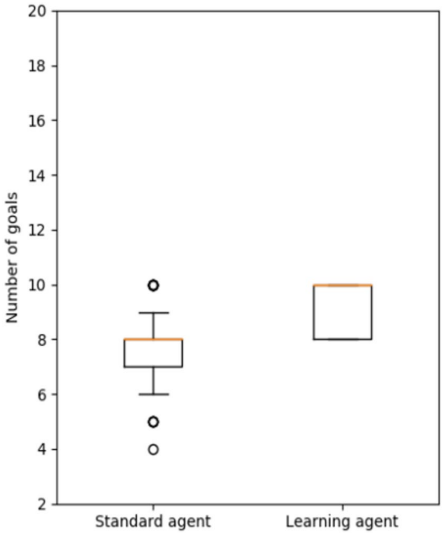

*Figure 5*. Experiment 1 box plots for standard and learning agents in 10-goal problems.

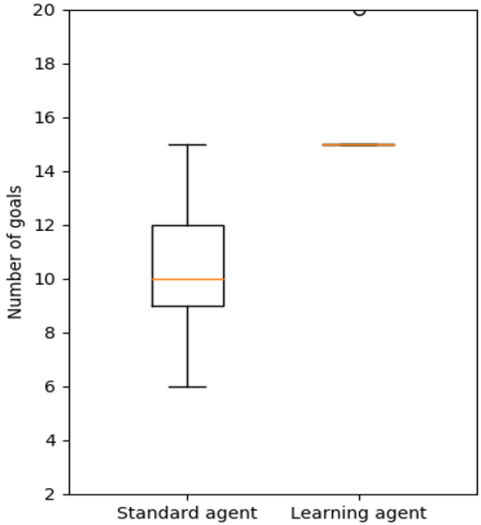

*Figure 6*. Experiment 1 box plots for standard and learning agents in 20-goal problems.

In Figures 5 and 6, the orange line represents the median of the distribution. Note that the y-axis reports the number of goals achieved instead of the percentage of goals achieved as in Figure 4. The black circles represent the outliers for the distribution of the data. In Figure 5, it is clear that the median of the standard agent is much less than the median of the learning agent. In Figure 6, the median of the learning agent is significantly greater than the median of the standard agent, and the variance is very low in contrast to the standard agent. By the time 20-goal problems are introduced, the learning agent has learned a complete operator. However, many cases will exist where goals cannot be achieved even with a perfect action model, so performance is less than 20 achievements out of 20 goals. For instance, consider again the example in Figure 3. Because herbicide cannot be sprayed outside the garden itself (e.g., at location (3,1)), the invasive at (3,2) cannot be killed without also killing the native plant.

In a second experiment with a 60:40 ratio of native to invasive plants, the learning agent still outperforms the standard agent but with less differences in performance (see Figure 7). The reason for this reduced outcome is that fewer examples of native plants being adjacent to invasive ones randomly occur within a 60:40 mix. Box plots are shown in Figures 8 and 9 to provide further details about the variance and significance of the results.

## 5. Related Research

As mentioned, computational metacognition has many interpretations in the literature. The cognitive psychology literature includes judgements of learning, feelings of knowing, tip of the tongue states and metamemory phenomena (Dunlosky & Thiede, 2013). But in some work, the definition is so broad as to blur the distinction between cognition and metacognition. For example, Kralik et al. (2018) defines metacognition as any decision process that takes input or output from another decision process. As such, they include planning as a metacognitive process as well as supervisory and arbitration processes. However, to argue that planning and simulation is metacognitive stretches the use of the term and provides little apparent value.

The recent work on computational approaches to metacognition are surprisingly thin. Most are either older papers or focus on human psychology rather than artificial intelligence. There is a paper on metacognition applied to neural networks (Babu & Suresh, 2012) and another applied to machine learning (Loeckx, 2017). Neither of these concern high-level reasoning and cognitive systems.

Similar to introspective monitoring and meta-level control, the Artificial Cognitive Neural Framework of Crowder, Friess, & Ncc (2011) splits computational metacognition into metacognitive experiences and metacognitive regulation. However, they also add a third component they call metacognitive knowledge or what a cognitive system knows about itself as a cognitive processor. The details of this framework are mainly conceptual, however, and no evaluation exists.

A few examples of metareasoning exist in the cognitive systems community such as the work of Martie, Alam, Zhang, & Anderson (2019). Their work on symbolic mirroring learns associations between neural network image classifiers and symbolic abstractions. They use meta-points to reflect on executed and unexecuted sections of the cognitive system itself by passing them to higher-level processes termed meta-operations. Although like MIDCA, the meta-operations examine and modify lower-level processes and involve explanation, much of the work is specific to vision tasks and less general.

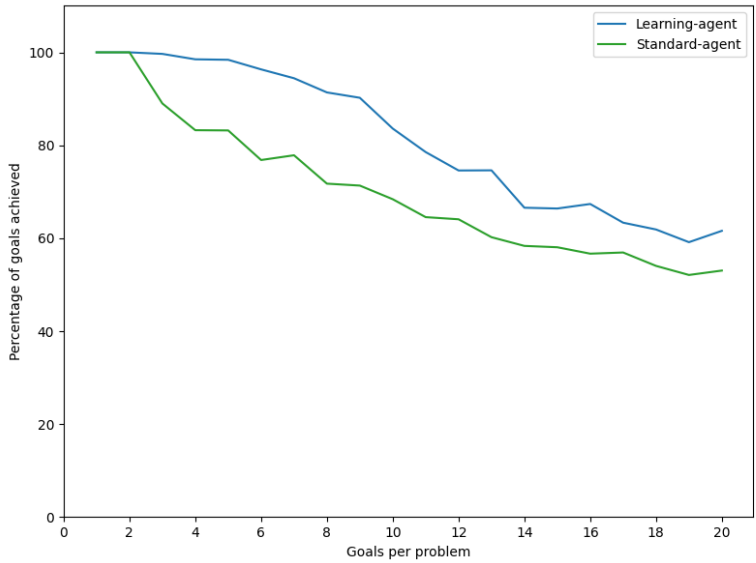

Figure 7. Experiment 2 performance as a function of problem complexity. The ratio of native to invasive plants is 60:40 in each of the 2,000 trials. As problems increase in complexity given ever more goals to achieve, performance goes down in terms of the percentage of goals achieved.

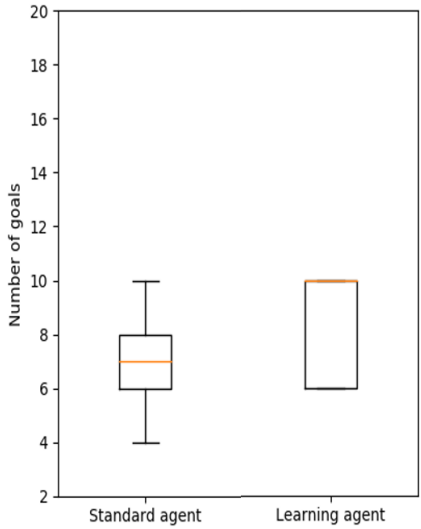 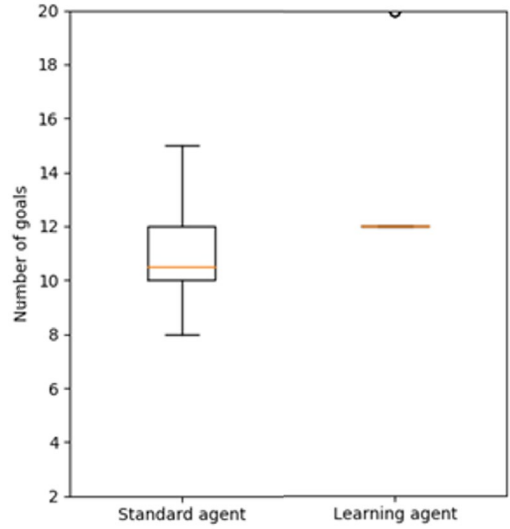

Figure 8. Experiment 2 box plots for standard and learning agents in 10-goal problems.

Figure 9. Experiment 2 box plots for standard and learning agents in 20-goal problems.

Human metacognition has been studied in the field of cognitive architectures: notably in ACT-R (Anderson, 2009; Larue, Hough, & Juvina, 2018), CLARION (Sun, 2016), and LIDA (Franklin et al., 2007). In ACT-R, Anderson and Fincham (2014) explored how reflective functions supported by metacognition can consciously assesses what one knows and how to extend it to solve a problem. More specifically, metacognition enabled the architecture to reflect on declarative representations of cognitive procedures, allowing for the modification or replacement of elements in the procedures used in mathematical problem-solving. More recent work explored metacognitive trigger mechanisms and more particularly the *feeling of rightness* (Wang & Thompson, 2019) metacognitive experience, and more particularly how feeling of rightness determines the depth of inner simulation of possible scenarios and hypothesis testing.

Metacognition is a key element of CLARION's hybrid (i.e., symbolic and subsymbolic) architecture. Following Flavell's (1976) original definition of metacognition, it supports the active monitoring and regulation of cognitive processes. Metacognitive mechanisms determine how much symbolic and sub-symbolic processing will be involved in decision making and can also dynamically change the ratio of symbolic and sub-symbolic processing and how they interact (e.g., which learning method(s) or reasoning mechanism(s) to apply). Internal parameters (e.g., learning rates, thresholds, and utilities) are also modified through metacognition.

In contrast to CLARION and MIDCA, LIDA has no specific metacognitive module. Higher-level cognitive processes are implemented as collections of behavior streams. Metacognitive processes specifically are a collection of behavior streams (i.e., sequences of actions) that control deliberation through internal actions (e.g., strategy regulation, resource allocation, and the interruption of strategies that have persisted too long in favor of an alternative). We have described metacognition as a kind of add-on to a cognitive system and implemented it as a separate software module in MIDCA. Yet this is not necessarily required computationally. Indeed, metacognition may be a self-reflective aspect of cognition itself. Only further research will clarify the implications of one approach as opposed to the other.

## 6. Conclusion

This paper briefly sketches a theory of computational metacognition analogous to a cognitive action-perception cycle and illustrates how it can be implemented in a cognitive architecture. We divide metacognition into three categories: explanatory, immediate and anticipatory metacognition. We discuss the first category in detail and briefly explain the last. Immediate metacognition remains for future research. Finally, we provide an empirical evaluation of explanatory metacognition and show how it improves system performance by progressively learning better action models. Previous publications (Cox & Dannenhauer, 2016; Cox, et al., 2017; Dannenhauer, et al., 2018) report similar results in alternative domains and therefore support the generality of the approach. However, this work is the first time the full metacognitive cycle in MIDCA has been described and demonstrated. Earlier publications focus on specific portions of the meta-level process such as the detection of metacognitive expectation failures, and they make large assumptions about other aspects of the meta-level computation.

Yet numerous limitations still exist with respect to the implementation and the theory. As the introduction admits, metacognition is not always the best choice for an agent or for humans. For example, if a system is under severe time constraints, reasoning about reasoning may waste valuable computational resources better spent on cognitive-level problem-solving or action execution. Although future research will provide a greater understanding of the trade-offs involved (see Norman, 2020, for some interesting heuristics that can potentially mitigate this problem), this paper clearly illustrates the positive effect computational metacognition can offer advanced cognitive systems.

Furthermore, some of the assumptions in this work remain implicit and underexplored. Although (in machines) statistical reinforcement learning (Sutton & Barto, 1999) or (in humans) operant conditioning (Skinner, 1938, 1957) may be at play at the lower levels of cognition, we view deliberate, *goal-driven learning* (Cox & Ram, 1999; Ram & Leake, 1995a) as a metacognitive activity. The agent needs to consider why reasoning fails and make decisions to achieve explicit goals to learn and hence (indirectly) improve performance. This perspective conflicts with standard formulations of the learning problem (e.g., Mitchell, 1997) even within the cognitive systems community (e.g., Langley, 2021; Mohan & Laird, 2014). As such, this makes the comparison to and contrast with other learning approaches difficult. Thus, an empirical comparison between an agent that learns with metacognition and one that learns without metacognition is problematic from our point of view. Again, future research will attempt to elaborate upon and evaluate this position and to clarify other assumptions less obvious in this paper.

Finally, as the examples shown here demonstrate, MIDCA implements relatively simple solutions for the problem of realizing computational metacognition. At this time, MIDCA's meta-level manages only singleton goal sets, and meta-level plans are currently very short and basic. Interactions between multiple meta-level goals and complex planning alternatives also remain for future work. But previous results shown at the cognitive level with MIDCA together with the robust state of the art in the planning community suggest that solutions to more complicated problems at the meta-level will be relatively straight forward and within practical reach. Rather than focus on sophisticated problems sets that rarely occur, we have instead put forth a fundamental approach to metacognition that promises to change the way we think about what makes a cognitive system intelligent and learning effective.

## Acknowledgements

This research is supported in part by the National Science Foundation through grant S&AS-1849131 and by the Office of Naval Research (ONR) through grant N00014-18-1-2009. We also thank the anonymous reviewers for their comments, insights, and suggestions and Matt Molineaux for suggesting the use of FOIL as a learning mechanism and his overall technical critique. Finally, we gratefully thank Paul Bello (a previous program manager at ONR) for taking a risk and funding the MIDCA project when it was merely a proposed concept ten years ago.

COMPUTATIONAL METACOGNITIONAnderson, J. R. (2009). *How can the human mind occur in the physical universe?* Oxford, UK: Oxford University Press.

Anderson, J. R., & Fincham, J.M. (2014). Extending problem-solving procedures through reflection. *Cognitive psychology* 74, 1–34.

Babu, G., & Suresh, S. (2012). Meta-cognitive neural network for classification problems in a sequential learning framework. *Neurocomputing* 81. 86-96. 10.1016/j.neucom.2011.12.001.

Boggs, J., Dannenhauer, D., Floyd, M. W., & Aha, D. W. (2018). The ideal rebellion: Maximizing task performance in rebel agents. In *Proceedings of the 6th Goal Reasoning Workshop*, held at IJCAI/FAIM-2018.

Brown, A. (1987). Metacognition, executive control, self-regulation, and other more mysterious mechanisms. In F. E. Weinert & R. H. Kluwe (Eds.), *Metacognition, motivation, and understanding* (pp. 65-116). Hillsdale, NJ: Lawrence Erlbaum Associates.

Conitzer, V. (2011). Metareasoning as a formal computational problem. In M. T. Cox & A. Raja (Eds.) *Metareasoning: Thinking about thinking* (pp. 121-127). Cambridge, MA: MIT Press.

Cox, M. T. (1997). Loose coupling of failure explanation and repair: Using learning goals to sequence learning methods. In D. B. Leake & E. Plaza (Eds.), *Case-Based Reasoning Research and Development: Second International Conference on Case-Based Reasoning* (pp. 425-434). Berlin: Springer-Verlag.

Cox, M. T. (2005). Metacognition in computation: A selected research review. *Artificial Intelligence* 169(2), 104-141.

Cox, M. T. (2011). Metareasoning, monitoring, and self-explanation. In M. T. Cox & A. Raja (Eds.) *Metareasoning: Thinking about thinking* (pp. 131-149). Cambridge, MA: MIT Press.

Cox, M. T. (2020). The problem with problems. In *Proceedings of the Eighth Annual Conference on Advances in Cognitive Systems*. Palo Alto, CA: Cognitive Systems Foundation.

Cox, M. T., Alavi, Z., Dannenhauer, D., Eyorokon, V., Munoz-Avila, H., & Perlis, D. (2016). MIDCA: A metacognitive, integrated dual cycle architecture for self regulated autonomy. In *Proceedings of the Thirtieth AAAI Conference on Artificial Intelligence*, Vol. 5 (pp. 3712-3718). Palo Alto, CA: AAAI Press.

Cox, M. T., & Dannenhauer, D. (2016). Goal transformation and goal reasoning. In *Proceedings of the 4th Workshop on Goal Reasoning*. New York, IJCAI-16.

Cox, M. T., & Dannenhauer, Z. A. (2017). Perceptual goal monitors for cognitive agents in changing environments. In *Proceedings of the Fifth Annual Conference on Advances in Cognitive Systems, Poster Collection* (pp. 1-16). Palo Alto, CA: Cognitive Systems Foundation.

Cox, M. T., Dannenhauer, D., & Kondrakunta, S. (2017). Goal operations for cognitive systems. In *Proceedings of the Thirty-first AAAI Conference on Artificial Intelligence* (pp. 4385-4391). Palo Alto, CA: AAAI Press.

Cox, M. T., Oates, T., & Perlis, D. (2011). Toward an integrated metacognitive architecture. In P. Langley (Ed.), *Advances in Cognitive Systems: Papers from the 2011 AAAI Fall Symposium* (pp. 74-81). Technical Report FS-11-01. Menlo Park, CA: AAAI Press.

Cox, M. T., & Ram, A. (1999). Introspective multistrategy learning: On the construction of learning strategies. *Artificial Intelligence,* 112, 1-55.
17